\documentclass{egpubl}
\usepackage{eg2024}
 
\ConferencePaper        
\CGFccby

\usepackage[T1]{fontenc}
\usepackage{dfadobe}  

\biberVersion
\BibtexOrBiblatex
\usepackage[backend=biber,bibstyle=EG,citestyle=alphabetic,backref=true]{biblatex} 
\electronicVersion
\PrintedOrElectronic

\ifpdf \usepackage[pdftex]{graphicx} \pdfcompresslevel=9
\else \usepackage[dvips]{graphicx} \fi

\usepackage{egweblnk} 

\usepackage{amssymb}
\usepackage{subcaption}
\usepackage{adjustbox}
\usepackage{multirow}
\usepackage{xspace}

\usepackage{lineno}

\title[Revitalizing Legacy Video Content: Deinterlacing with Bidirectional Information Propagation]%
      {Revitalizing Legacy Video Content: \\ Deinterlacing with Bidirectional Information Propagation}

\author[Gao et al.]
{\parbox{\textwidth}{
    \centering 
    Zhaowei Gao$^{*1,2}$,
    Mingyang Song$^{*1,2}$,
    Christopher Schroers$^{2}$ and 
    Yang Zhang$^{2}$
} \\
{\parbox{\textwidth}{
    \centering 
    $^1$ETH Zurich, Switzerland\\
    $^2$DisneyResearch|Studio, Switzerland \\
    $ $ \texttt{\{zhagao, misong\}@student.ethz.ch, \{christopher.schroers, yang.zhang\}@disneyresearch.com}
}}
}

\begin{document}
 \teaser{
  \includegraphics[width=\linewidth]{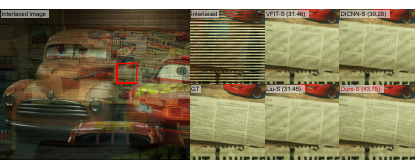}
  \centering
   \caption{\emph{Left}: Visualization of interlaced content.
   \emph{Right}: Close ups of the interlaced input, the ground truth target, as well as the output of our small model versus existing methods at the same parameter level (0.5M). The PSNR values, written in the brackets,
   are computed on the cropped region. 
   For more visual details, please refer to the Sec. \ref{sec:visual result}.}
 \label{fig:teaser}
}

\maketitle
\begin{abstract}
Due to old CRT display technology and limited transmission bandwidth, early film and TV broadcasts commonly used interlaced scanning. This meant each field contained only half of the information. Since modern displays require full frames, this has spurred research into deinterlacing, i.e. restoring the missing information in legacy video content. In this paper, we present a deep-learning-based method for deinterlacing animated and live-action content. Our proposed method supports bidirectional spatio-temporal information propagation across multiple scales to leverage information in both space and time. 
More specifically, we design a Flow-guided Refinement Block (FRB)
which performs feature refinement including alignment, fusion, and rectification. Additionally, our method can process multiple fields simultaneously, reducing per-frame processing time, and potentially enabling real-time processing. Our experimental results demonstrate that our proposed method achieves superior performance compared to existing methods.


\end{abstract}  
\def\thefootnote{*}\footnotetext{These authors contributed equally to this work}\def\thefootnote{\arabic{footnote}}
\section{Introduction}

\begin{figure}[t]
  \centering
  \includegraphics[width=0.45\textwidth]{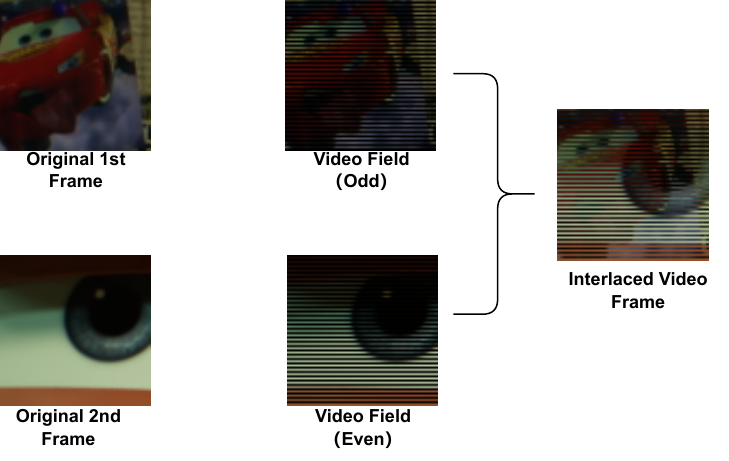}
  \caption{Illustration of the interlacing process. Two consecutive original frames are combined
into a single interlaced frame by selecting odd-numbered rows from the first frame and even-numbered rows from the second frame.
}
  \label{fig:k_surf}
\end{figure}

Interlaced video was developed in the early days of television to balance visual quality and technical constraints within limited bandwidth and refresh rates. It captured odd and even fields in alternating frames, combining them into interlaced frames for displaying on screens, as illustrated in Figure~\ref{fig:k_surf}. While interlacing was once a useful technique, modern displays require progressive video, making interlaced formats obsolete. However, in the past, when interlacing videos, the original frames usually were not preserved. Consequently, deinterlacing has become crucial for the restoration of old interlaced content.

Deinterlacing involves estimating the content of absent lines within each field of an interlaced video signal, aiming to generate the complete frame information while ensuring visual quality and minimizing artifacts. A large variety of deinterlacing algorithms exists: Conventional Deinterlacing methods\cite{705528,6854719,wang2012efficient} can be categorized into intra-field interpolation, inter-field interpolation, and motion-based. Intra-field interpolation reconstructs the missing field by averaging pixel values available from the current field. Inter-field interpolation replicates information from neighboring fields to approximate the missing field. The outcome of such methods is generally unsatisfactory due to the simplicity of replicating and averaging pixel values. Despite involving motion detection and alignment, conventional motion-based methods are still insufficient in capturing accurate inter-frame correspondences. Fortunately, deinterlacing is perfectly suited for fully supervised training since the degradation process induced through interlacing is well-defined. This allowed to harness the expressive power of neural networks and to significantly surpass the previously available handcrafted reconstruction strategies across diverse input data%
\cite{zhao2021rethinking, zhu2017realtime, liu2021spatial, yeh2022vdnet}.



Sharing a similar goal of restoring missing information from observations, video super resolution\cite{chan2021basicvsr,xiang2020zooming,chan2022basicvsr++}, video frame interpolation\cite{shi2022video}, as well as image and video restoration \cite{liang2022vrt, chen2022simple,wang2019edvr} can offer valuable insights for video deinterlacing, especially when it comes to devising strategies for temporal propagation, alignment and fusion.

In order to make the most effective use of both spatial and temporal information in interlaced videos, we propose a Flow-guided Refinement Block (FRB). Opposed to \cite{chan2022basicvsr++}, we introduce an additional fusion mechanism after the deformable convolutions.
While \cite{chan2022basicvsr++} employs recurrent propagation, we leverage bidirectional parallel propagation~\cite{liang2022vrt} on each scale level. 
Our framework processes six consecutive fields from interlaced frames at once and predicts the six corresponding missing fields.


The main contributions of our work are:
\begin{itemize}
    \item We propose a deep learning framework for deinterlacing that incorporates a mechanism for the propagation of temporal information in both image and latent space, as well as feature refinement. Our framework effectively tackles the restoration of interlacing artifacts, including combing and aliasing.
    \item Our model is lightweight and capable of simultaneously outputting six deinterlaced video frames. This makes it a promising candidate for real-time deinterlacing applications.
    \item Our extensive experimental results demonstrate that our proposed method can remove complex interlacing artifacts and achieve state-of-the-art performance.
\end{itemize}


\section{Related Work}
\label{sec:related_work}

\subsection{Deinterlacing Techniques}\label{deinterlacing}
\subsubsection{Conventional Deinterlacing}
Image and video deinterlacing represent classic challenges in the field of computer vision. Existing conventional methods can be categorized into three primary groups: intra-field deinterlacing, inter-field deinterlacing and motion-based adaptation and compensation. Intra-field deinterlacing techniques independently reconstruct two complete frames from the odd and even fields. However, the previous methods were simply calculated the average of the lines, that immediately above and below the missing line, leading to lower visual quality. Due to the fact that most interlacing artifacts appear around the edge, subsequent work has placed greater emphasis on edge area to improve the edge line average\cite{1998}. Other approaches like bilateral filtering model\cite{wang16}, locality and similarity adaption\cite{wang2012efficient}, and moving least square methods\cite{6469202} have further improved the results of the removing interlaced artifacts in edge area. While these techniques can perform the deinterlacing of frames and generate the missing components, their performance remains suboptimal. In contrast, inter-field deinterlacing methods\cite{6490258,4801598} aims to enhance visual quality by incorporating temporal information from neighboring fields and multiple fields during frame reconstruction. However they mainly just replicate the weighted content from the preceding field, and the outcome is usually unsatisfactory.
Motion-based adaptation and compensation
methods\cite{mohammadi2012enhanced, 1205477}typically require accurate motion compensation or motion estimation to achieve satisfactory quality, which can be a challenging task for conventional deinterlacing methods. Hence, when large motion exists between these frames, visual artifacts become apparent.

\subsubsection{Deep Learning-based Deinterlacing}

In the above section, we have already mentioned that due to the well-known and explicit degradation process of interlacing, it seems that deinterlacing is perfectly suited for fully supervised training and an ideal candidate for a deep learning based solution. With the advancement of deep learning technology, an increasing number of deinterlacing networks based on deep learning have emerged. In 2017, Zhu et al.\cite{zhu2017realtime} introduced the first Deep Convolutional Neural Network approach(DICNN)for deinterlacing, emphasizing real-time processing to achieve a balance between speed and quality. Further, Liu et al.\cite{liu2021spatial} devised a neural deinterlacing network using deformable convolution and attention-residual blocks. Zhao et al.\cite{zhao2021rethinking} used a two-stage deinterlacing ResNet Structure to deal with complex interlacing artifacts. Yet, these approaches consider only intra-frame deinterlacing without fully leveraging the temporal information. In \cite{bernasconi2020deep}, Bernasconi et al. presented a multi-field deinterlacing method based on residual dense network. Recently, VDNet\cite{yeh2022vdnet} has proposed an RNN-based deinterlacing framework that leverages deformable blocks to align feature between different frames.
However their feature-domain alignment of supporting fields is suboptimal and still cannot handle the complicated artifacts in the presence of large motion. 


\subsection{Spatio-temporal Upscaling}\label{sr}

Video deinterlacing can also be considered a type of video upscaling task, where a field can be seen as an image that needs vertical upscaling by a factor of 2. Existing methods for video upscaling
typically rely on optical flow estimation to warp supporting frames to align with a reference frame, as seen in \cite{caballero2017real}. However, accurate flow estimation and warping can be challenging and introduce artifacts. TDAN\cite{tian2020tdan}, EDVR\cite{wang2019edvr} and VFIT\cite{shi2022video} offer alternative approaches that eliminate the need for motion estimation, using deformable convolution for alignment. 
BasicVSR++\cite{chan2022basicvsr++} further improves the approach of its predecessor, BasicVSR\cite{chan2021basicvsr}, with second-order grid propagation and flow-guided deformable convolution. In TMNet \cite{xu2021temporal} proposed using bi-directional Deformable ConvLSTM for spatio-temporal upscaling. 

\section{Method}

In this section, we will outline how we pre-process the images in Sec. \ref{sec:data_pre_process}. Subsequently, we will present our proposed deinterlacing architecture in detail in Sec: \ref{sec:proposed_method}.

\subsection{Data processing pipeline}
\label{sec:data_pre_process}
\begin{figure}[t]
  \centering
  \includegraphics[width=0.48\textwidth]{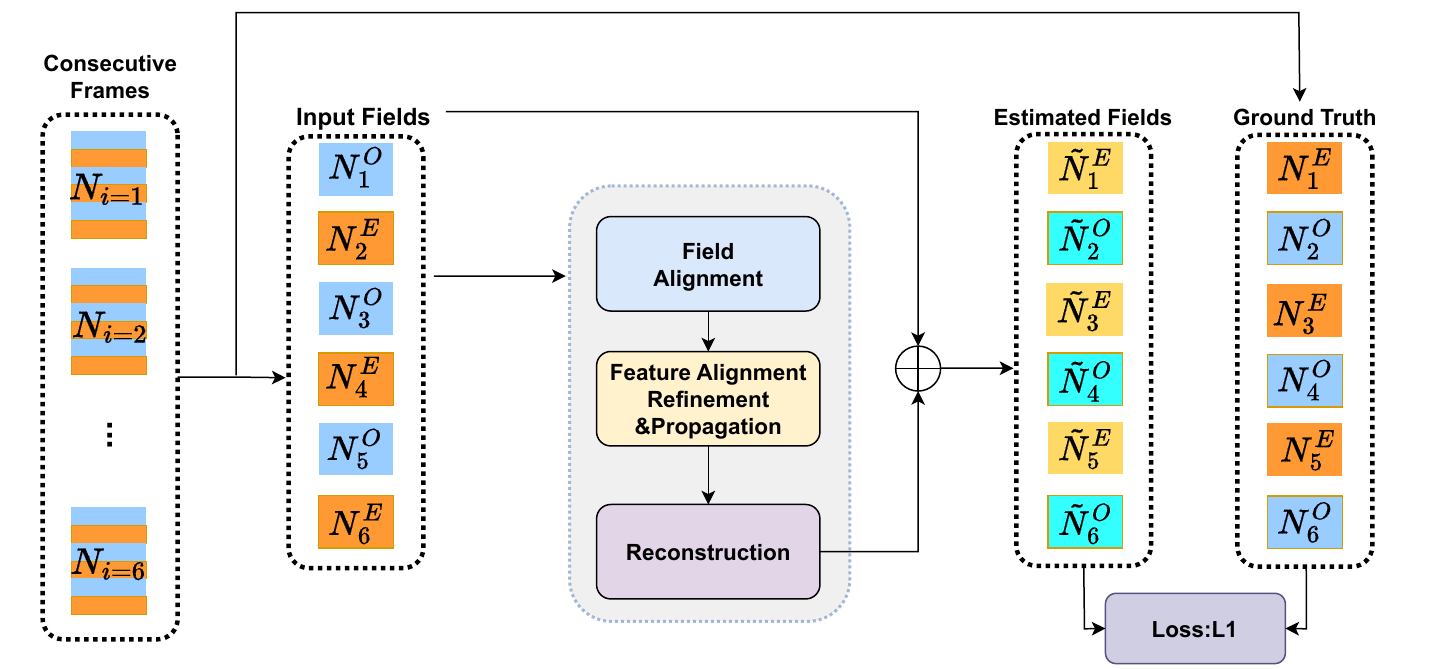}
  \caption{Overview of data processing during training. We utilize sequences of six consecutive frames, splitting each frame into odd and even fields. The input order for the fields is odd from the first frame, even from the second frame, and then alternates between odd and even for the subsequent frames. The output $\tilde{N}_i^{O}$, $\tilde{N}_i^{E}$ is an estimation of the missing half information. $N_i$ represents the original t-th frame. $N_i^{O}$ and $N_i^{E}$ represent \textit{odd} and \textit{even} fields of frame $t$}
  \label{fig:data}
\end{figure}

Our data processing pipeline is depicted in Fig. \ref{fig:data}. We sample the odd or even field alternatively from 6 consecutive frames as the input to our model. The model predicts the rest of the corresponding even and odd fields and calculates the objective error during the training process.


Specifically, the order of the input fields follows the role where the first field ($N_1^O$) from the odd-field of the first frame, then the second field ($N_2^E$) from the even-field from the second frame, and then alternates between odd and even for the subsequent input fields. The output is an estimation of the missing half-frame information, where the output order for the fields is even-field ($\tilde{N}_1^E$) for the first frame, odd-field ($\tilde{N}_2^O$) for the second frame, and then alternates between odd- and even-field for the subsequent frames. It is worth noting that our model outputs six fields in a single forward pass, which helps reduce the processing time per video frame and accelerates the overall processing speed. This provides potential capabilities for real-time deinterlacing.

\subsection{The proposed method}
\label{sec:proposed_method}

\begin{figure*}[ht]
  \centering
  \includegraphics[width=1\textwidth]{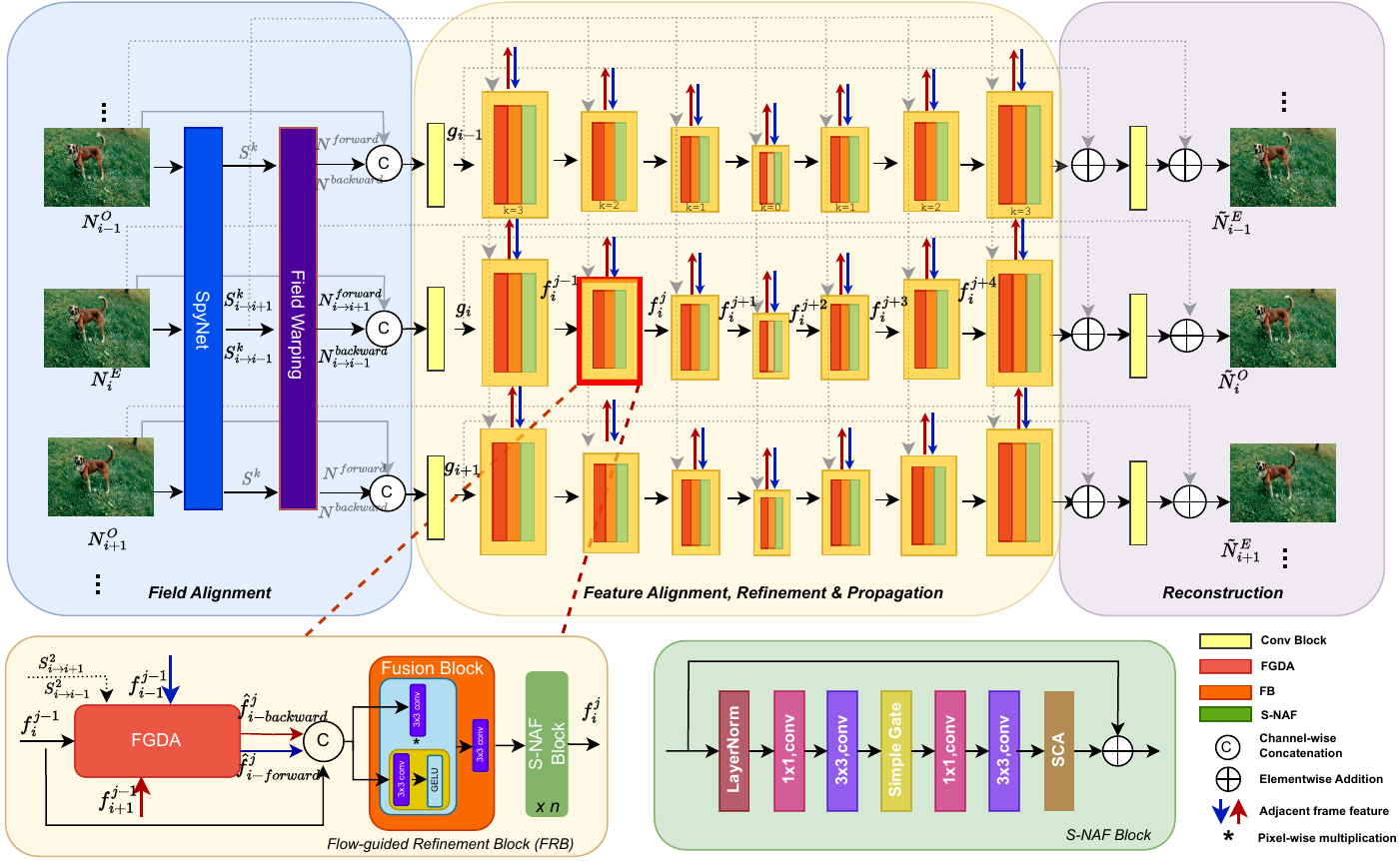}
  \caption{
  Overview of our deinterlacing network. We introduce forward-backward propagation to refine features bidirectionally. Specifically, within each propagation block, we introduce a Flow-guided Refinement Block (FRB). In the FRB, the FGDA block was designed to enhance offset diversity for the deformable convolution. It is followed by a Fusion block and S-NAF block to further refine the aligned features.
  }
  \label{fig:pro}
\end{figure*}

As mentioned in Sec. \ref{sec:related_work}, video processing tasks often benefit from the utilization of temporal information, however, it is also challenging. The difficulty lies in the need to aggregate information between multiple correlated frames in a video sequence that contains complex moving objects. Therefore, alignment and propagation of temporal sequence information become crucial.

The proposed overall architecture is shown in Fig~\ref{fig:pro}. The alignment in our proposed method can be categorized into image space alignment and feature space alignment. Feature alignment leverages a UNet-like structure and aligns at different scales. Building on the concept of BasicVSR++\cite{chan2022basicvsr++}, we propose a Flow-guided Refinement Block (FRB). It integrates Flow-guided Deformable Alignment (FGDA) and a Fusion Block (FB) in conjunction with SimpleNAF (S-NAF) blocks. This helps to overcome instability during the training of Deformable Convolution Network (DCN), which can suffer from overflow issues.

For information propagation, the commonly used unidirectional propagation transmits information from the first frame to the next in the video sequence. However, the information received by different frames is unbalanced. Specifically, the first frame receives no information from the video sequence except itself, whereas the last frame receives information from all the previous frames. Therefore, the later frames receive more information than earlier frames, which may result in sub-optimal outcomes and produce temporal artifacts, such as quality fluctuation over time. To address this, we developed a bidirectional information propagation scheme. 

As shown in Fig~\ref{fig:pro}, given an input of six consecutive fields, SPyNet\cite{ranjan2017optical} is first applied to estimate optical flow, $S_i^k$, between each pair of neighboring fields, followed by a forward and backward alignment of adjacent fields in the image domain, $N_i^{forward}$ and $N_i^{backward}$. Then the warped fields are concatenated with the input fields along the channel dimension. After that, a 3D convolutional layer is applied to extract features ($g_i$) from each field and warped field. In the Feature Alignment, Refinement, and Propagation (FARP) component, $f_i^j$ from each Flow-guide Refinement Block (FRB) is then propagated under our bidirectional propagation scheme across corresponding scales, where alignment is performed by our FGDA module and feature refinement is conducted by the FB and S-NAF modules. After propagation, the aggregated features are used to reconstruct the output image through convolutional layers.
 
In the following sections, a detailed description of the three components of our model will be presented respectively, including Field Alignment, Feature Alignment, Refinement $\&$ Propagation (FARP), and Reconstruction.

\subsubsection{Field Alignment}

We first perform alignment in the image domain. Alignment is achieved by utilizing a pre-trained SPyNet\cite{ranjan2017optical} to compute optical flow followed by forward and backward warping. It's worth noting that we apply spatial alignment at four different scales with corresponding optical flow. After warping and upsampling to the original scale, the original image fields $N_{i}$, and four pairs of $N^{forward}$ and $N^{backward}$ are concatenated along the channel dimension. Moreover, the four different scales of optical flows have been further utilized as inputs to the subsequent FRBs at various scales accordingly in the FARP component.


\subsubsection{Feature Alignment, Refinement $\&$ Propagation (FARP)}

\begin{figure}[t]
  \centering
  \includegraphics[width=0.5\textwidth]{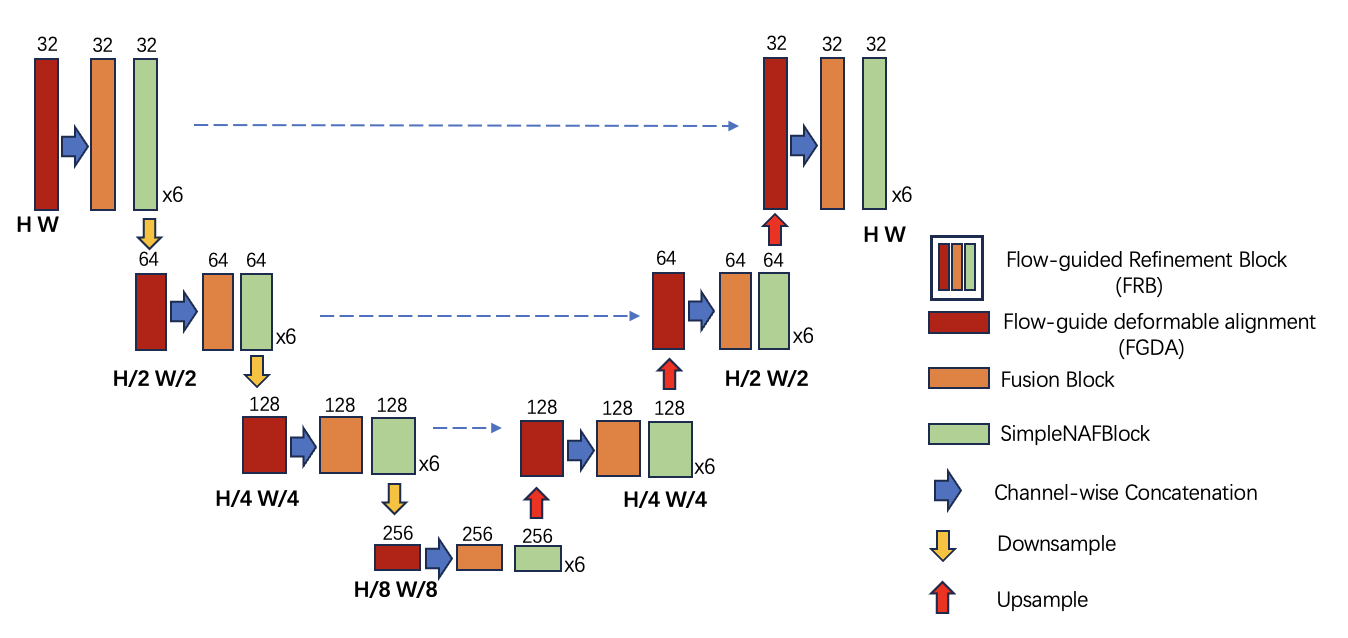}
  \caption{Illustration of the proposed UNet-like architecture of the FARP. The inputs to each scale consist of the corresponding optical flow $S_{i}^{k}$, the feature $f_i$ from the current field, and the features $f_{i-1}$, $f_{i+1}$ from the previous and next fields.}
  \label{fig:U}
\end{figure}

We develop a bidirectional UNet-like scheme to facilitate refinement through propagation where the intermediate features are initially propagated independently both forward and backward in time and then down- and up-sampled and finally formed the aggregation process. The graphical illustration of the UNet-like structure is shown in Fig~\ref{fig:U}. Through this refinement process, the receptive field can be expanded and the information from different frames can be `revisited' and employed for feature enhancement.

Specifically, after the field alignment, a 3D convolutional layer is applied to extract image features from the input. The features are then propagated under our bidirectional UNet-like propagation scheme in latent space, where alignment and refinement are performed in the feature domain under four various scales by our Flow-guided Refinement Block (FRB), as shown in Fig. \ref{fig:pro}.

In the following subsections, we provide a detailed explanation of the forward feature propagation in our proposed FRB module. The process for backward propagation is similarly defined. 

\textbf{Flow-guided Refinement Block (FRB)} As shown in Fig. \ref{fig:pro}, let $N_{i}$ be the input image, $g_{i}$ be the feature extracted from the convolutional layer. $f_{i}^{j}$ be the feature computed at the $i$-th timestep in the $j$-th propagation block. To compute the forward and backward feature of $f_{i}^{j}$, we first align $f_{i+1}^{j-1}$ and $f_{i-1}^{j-1}$ using the flow-guided deformable alignment (FGDA) module, respectively.
\begin{equation}
  \hat{f}_{i-forward}^{j} =  \textrm{FGDA}\left ( f_{i}^{j-1}, f_{i+1}^{j-1}, S_{{i \to i+1}}^{k} \right )    
\end{equation}
\begin{equation}
  \hat{f}_{i-backward}^{j} =  \textrm{FGDA}\left ( f_{i}^{j-1}, f_{i-1}^{j-1}, S_{{i \to i-1}}^{k} \right )   
\end{equation}
where $ S_{{i \to i+1}}^{k}$ , $ S_{{i+1 \to i}}^{k} $ denote the optical flows at $k$-th scales from $i$-th field to the $(i+1)$-th and $(i-1)$-th field, respectively. And $f_{i}^{0} = g_{i}$. The features from the current scale and from corresponding scales of adjacent fields are then concatenated and aggregated by an FB and then passed through multiple S-NAF blocks for further refinement. The S-NAF block was proposed in \cite{song2023generative} and can make model architecture simpler and leaner. This operation can be formulated as below:
\begin{equation}
  f_{i}^{j} =  \textrm{S-NAF}\left ( \textrm{FB}\left(\mathbb{C}\left (f_{i}^{j-1}, \hat{f}_{i-forward}^{j}, \hat{f}_{i-backward}^{j} \right )\right ) \right)    
\end{equation}
where $\mathbb{C}$ denotes concatenation along channel dimension.



\begin{figure}[t]
  \centering
  \includegraphics[width=0.47\textwidth]{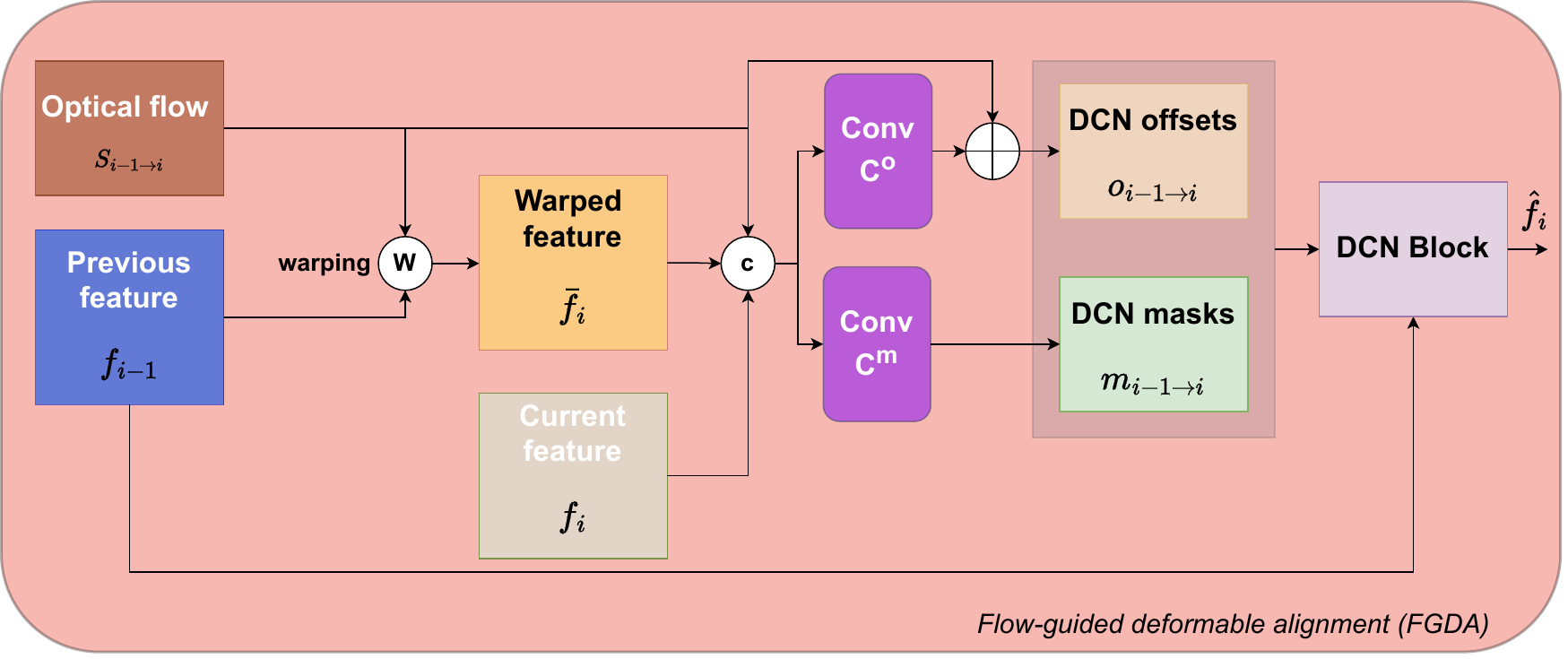}
 \caption{Illustration of the Flow-guided deformable alignment (FGDA) module proposed by~\cite{chan2022basicvsr++}. The optical flow at corresponding scales is used for feature warping. The warped feature, the current feature and the optical flow are then concatenated to produce DCN offsets and masks. A DCN is then applied to the unwarped feature for feature alignment. Only forward propagation is shown in this figure, backward propagation is omitted for simplicity.}
  \label{fig:fdcn}
\end{figure}

\textbf{Flow-Guided Deformable Alignment (FGDA)} As an essential component of our work, we explain the design of FGDA in BasicVSR++~\cite{chan2022basicvsr++} for self-containing. Whereas the deformable alignment has achieved better performance over flow-based alignment, thanks to the offset diversity inherently introduced in deformable convolution (DCN)\cite{dai2017deformable}, the instability in vanilla DCN could lead to offset overflow, thus reducing final performance.
Given the strong relation between the deformable alignment and flow-based alignment, optical flow is utilized to further guide deformable alignment, in order to fully utilize offset diversity and address the instability issue. The FGDA module has been illustrated in Fig. \ref{fig:fdcn}, we omit the superscript $j$ and $k$ in the notation, and only forward propagation has been demonstrated for simplicity.

Specifically, in Fig. \ref{fig:fdcn}, the current feature $f_{i}$ at timestep $i$, the feature $f_{i-1}$ computed from timestep $i-1$, and the optical flow $ S_{{i-1 \to i}}$ to the current field are the inputs. Firstly, $f_{i-1}$ is forward warpped with $ S_{{i-1 \to i}}$:
\begin{equation}
    \bar{f_{i}}=\mathcal{W}\left (f_{i-1}, S_{{i-1 \to i}}  \right )    
\end{equation}
where $\mathcal{W}$ represents the spatial warping operation. The aligned features $\bar{f_{i}}$ are subsequently employed to calculate the DCN offsets $o_{{i-1 \to i}}$ and modulation masks $m_{{i-1 \to i}}$. Rather than directly computing the DCN offsets, the residue with respect to the optical flow is computed by $\textrm{Conv}^O$:
\begin{equation}
    o_{{i-1 \to i}}=S_{{i-1 \to i}} + \textrm{Conv}^{O}\left (\mathbb{C}(f_{i}, \bar{f_{i}}, S_{{i-1 \to i}})   \right ) 
\end{equation}
\begin{equation}
    m_{{i-1 \to i}}= \sigma \left ( \textrm{Conv}^{M}\left (\mathbb{C}(f_{i}, \bar{f_{i}}, S_{{i-1 \to i}} )   \right ) \right)
\end{equation}
where $\textrm{Conv}^{O, M}$ represents a stack of convolutional layers for ${o}$ and $m$ prediction respectively. They share a similar architecture and are detailed in Table ~\ref{tab:1}. $\sigma $ denotes the sigmoid activation function. Subsequently, output feature $\hat{f}_{i}$ can be obtained by a DCN block with the input of feature $f_{i-1}$, offset $o_{{i-1 \to i}}$ and mask $m_{{i-1 \to i}}$. 
\begin{equation}
 \hat{f}_{i}=\textrm{DCN}\left(f_{i-1}, o_{{i-1 \to i}}, m_{{i-1 \to i}} \right)
\end{equation}

The aforementioned formulations can be used for the forward propagation of a single field feature. The same process can be independently applied for backward propagation.

\subsubsection{Reconstruction }
As shown in Fig.~\ref{fig:pro}, after obtaining the aggregated and refined features spatially and temporally in the FARP component, a 3D convolutional layer is employed to reconstruct the color information of the predicted image from the latent space. Additionally, skip connections are applied both in the latent space and the image space as a residual process. This residual mechanism is responsible for obtaining the signal to refine the final output of the deinterlaced image. Meanwhile, this also allows gradients to propagate more easily back to earlier layers, enabling the model to learn more complex features without suffering from gradient vanishing issues and enhancing the quality of the deinterlaced image.

\section{Experiments}
\subsection{Implement detail}
\label{sec:implement detail}
We use a pretrained SpyNet\cite{ranjan2017optical} for optical flow estimation. Due to the pyramid structure of SpyNet, we obtain flow at 4 different scales. We have designed two networks with different amounts of parameters, namely \textit{Ours-S} contains 0.5M and \textit{Ours-L} contains 9M parameters. The number of FRBs was set to 7. The number of S-NAF blocks for each FRB was set to 3 and 6 for Ours-S and Ours-L, respectively. The hyperparameter of the \textit{dim} in Table. \ref{tab:1} was designed for the feature input channel in each FRB as follows, where \textit{dim} = [20, 20, 20, 20, 20, 20, 20] and \textit{dim} = [20, 40, 80, 160, 80, 40, 20] are applied for Ours-S and Ours-L model. The DCN kernel size was set to 3 and the number of deformable groups was set to 4.

\begin{table}[t]
\centering
\begin{tabular}{|c|c|c|}
\hline
\hline
Layer & \multicolumn{1}{c|}{$\textrm{Conv}^{O}$}  &  \multicolumn{1}{c|}{$\textrm{Conv}^{M}$} \\
\hline
1. &  \multicolumn{2}{c|}{conv(dim*2+2, dim, 3)} \\
\hline
2. &  \multicolumn{2}{c|}{LeakyReLU(0.1)}\\
\hline
3. &  \multicolumn{2}{c|}{conv(dim, dim, 3)} \\
\hline
4. &  \multicolumn{2}{c|}{LeakyReLU(0.1)}\\
\hline
5. &  \multicolumn{2}{c|}{conv(dim, dim, 3)} \\
\hline
6. &  \multicolumn{2}{c|}{LeakyReLU(0.1)}\\
\hline
7.  & \multicolumn{1}{c|}{conv(dim,288,3)}  &  \multicolumn{1}{c|}{conv(dim,144,3)} \\
\hline
\hline
\end{tabular}
\caption{The architecture of $\textrm{Conv}^{O, M}$. More detailed information regarding the list \textit{dim} can be found in Sec.~\ref{sec:implement detail}.\\}
\label{tab:1}
\end{table}

\subsection{Training and Testing Datasets}
We utilize datasets consisting of natural video sequences and synthesize the interlaced frames for both training and evaluation with the method mentioned in Sec. \ref{sec:data_pre_process}. We trained our models with the Vimeo-90K~\cite{xue2019video} training set that contains 64,612 sequences and tested our models on the remaining 7,824 testing sequences. To assess the generalization capability of our model across diverse data distribution, we utilized Vid4\cite{5995614}, SPMC\cite{Tao_2017_ICCV}, and UDM10\cite{yi2019progressive} for additional testing without retraining or fine-tuning our models.


\subsection{Training Setting}
We adopt AdamW~\cite{loshchilov2017decoupled} optimizer, and the learning rate decays from $1\times10^{-4}$ to $1\times10{-7}$ with Cosine Annealing~\cite{loshchilov2016sgdr} scheduler. The training process consists of 600K iterations. The batch size was set to 8 and the patch size was $128\times128$. 
Our models were end-to-end trained via a $L_{1}$ loss function. All the experiments were performed on one Nvidia GeForce RTX 3090.

\subsection{Evaluation Metrics}
To conduct a comprehensive evaluation, we compare our approach to previous methods in terms of restoration accuracy and inference efficiency. In order to fairly compare with existing methods, we followed the evaluation method in \cite{yeh2022vdnet}. To assess the fidelity, we employ Peak Signal-to-Noise Ratio (PSNR) and Structural Similarity Index Measure (SSIM)~\cite{1284395} as evaluation metrics. The reported scores were derived by calculating the average scores across the entire test set. 
As for efficiency evaluation, we calculate the runtime based on an image crop with 256 $\times$256 resolution on various models with similar amounts of parameters.



\subsection{Comparisons to existing methods}

\begin{table*}[ht]
  \centering
  \begin{tabular}{l c c c c c c }
    \hline
    \hline

    \multirow{2}{*}{\textbf{Method}} &\textbf{Parameters} & \textbf{Runtime} &  \textbf{VimeoTest} & \textbf{Vid4} & \textbf{SPMC} & \textbf{UDM10} \\
        \cline{4-4} \cline{5-5} \cline{6-6} \cline{7-7}
                            & (Million) &  (ms)  & PSNR \quad SSIM & PSNR\quad SSIM & PSNR\quad SSIM & PSNR\quad SSIM \\
    \hline
    Liu-S\cite{liu2021spatial} & 0.52 & 169.33 & 40.45 \quad  0.9804 & 31.24 \quad  0.9524 & 36.73 \quad  0.9740 & 42.12 \quad  \textcolor{blue}{\underline{0.9872}} \\

    VFIT-S\cite{shi2022video} & 0.51 & 46.85 & 40.79 \quad 0.9824 & 31.30 \quad 0.9541 & 40.89 \quad 0.9882 & 41.06 \quad 0.9836 \\

    DICNN-S\cite{zhu2017realtime} & 0.54 & \textcolor{blue}{\underline{20.35}} & 41.42 \quad0.9831 & 31.77 \quad0.9559 & 40.58 \quad0.9881&41.58\quad 0.9844 \\

    VDNet-S$^{\dagger}$\cite{yeh2022vdnet} &  0.51 & \textbf{-} & \textcolor{blue}{\underline{42.68}} \quad \textcolor{blue}{\underline{0.9848}} & \textcolor{blue}{\underline{32.26}} \quad \textcolor{blue}{\underline{0.9568}} & \textcolor{blue}{\underline{43.17}} \quad \textcolor{blue}{\underline{0.9907}}& \textcolor{blue}{\underline{42.48}} \quad 0.9865\\

    Ours-S & 0.50  & \textcolor{red}{\underline{18.45}} & \textcolor{red}{\underline{44.40}} \quad \textcolor{red}{\underline{0.9906}} & \textcolor{red}{\underline{34.20}} \quad\textcolor{red}{\underline{0.9703}} & \textcolor{red}{\underline{46.35}} \quad\textcolor{red}{\underline{0.9959}}& \textcolor{red}{\underline{44.49}} \quad \textcolor{red}{\underline{0.9914}}\\
    \hline
    \hline
    
    Liu-L\cite{liu2021spatial} & 9.12 & 1593.99 & 40.70 \quad  0.9810 & 30.61 \quad  0.9498 & 36.99 \quad  0.9749 & 42.27 \quad  0.9875 \\

    VFIT-L\cite{shi2022video} & 8.87 & \textcolor{blue}{\underline{87.13}} & 43.75 \quad 0.9891 & 34.07 \quad 0.9696 & 45.27 \quad 0.9945 & 43.51 \quad 0.9898 \\

    DICNN-L\cite{zhu2017realtime} & \textbf{-} & \textbf{-}  & \textbf{-} \quad\quad\quad \textbf{-} &  \textbf{-}\quad\quad\quad \textbf{-} & \textbf{-} \quad\quad\quad \textbf{-}  & \textbf{-}\quad\quad\quad \textbf{-}\\

    TMNet$^{\dagger}$\cite{xu2021temporal} & 12.44 & \textbf{-}  & 45.70 \quad0.9910 & 34.53\quad0.9698 &47.26 \quad0.9958  &44.59\quad0.9912\\  

    VDNet-L$^{\dagger}$\cite{yeh2022vdnet} &  9.23 & \textbf{-} & \textcolor{blue}{\underline{46.45}} \quad \textcolor{blue}{\underline{0.9922}} & \textcolor{blue}{\underline{34.83}}\quad \textcolor{blue}{\underline{0.9703}} & \textcolor{blue}{\underline{47.84}} \quad \textcolor{blue}{\underline{0.9965}}& \textcolor{blue}{\underline{45.52}} \quad \textcolor{blue}{\underline{0.9928}}\\

    Ours-L & 8.88  & \textcolor{red}{\underline{26.34}} & \textcolor{red}{\underline{46.50}} \quad \textcolor{red}{\underline{0.9935}} & \textcolor{red}{\underline{35.46}}\quad\textcolor{red}{\underline{0.9749}} & \textcolor{red}{\underline{48.19}} \quad\textcolor{red}{\underline{0.9972}}& \textcolor{red}{\underline{46.20}} \quad \textcolor{red}{\underline{0.9940}}\\
    \hline
    \hline
  \end{tabular}
 \caption{Quantitative comparison (PSNR/SSIM). \textcolor{red}{Red} and \textcolor{blue}{blue} colors represent the best and second-best performance, respectively. We reimplement and train the models Liu\cite{liu2021spatial}, VFIT\cite{shi2022video}, DICNN\cite{zhu2017realtime} on the VimeoTrain\cite{xue2019video} dataset at two distinct parameter levels (Large:9M, Small:0.5M). Notably, without further retraining or fine-tuning on the VimeoTest\cite{xue2019video}, Vid4\cite{5995614}, SPMC\cite{Tao_2017_ICCV}, and UDM10\cite{yi2019progressive} datasets, both of our models consistently achieve superior performance at a shorter runtime compared to the other methods. The runtime is calculated based on an image size of 256×256. The remaining empty cells indicate results that were not reported in previous studies. $^{\dagger}$: numbers are taken from \cite{yeh2022vdnet}. }
 \label{tab:2}
\end{table*}

We compared our proposed method with existing deinterlacing and video frame interpolation methods. For the deinterlacing methods, we compared ours with VDNet\cite{yeh2022vdnet}, Liu\cite{liu2021spatial} and DICNN\cite{zhu2017realtime}. For the video spatio-temporal upscaling methods, we choose the SOTA method TMNet\cite{xu2021temporal} and VFIT\cite{shi2022video} as the benchmark. We re-implemented the models Liu\cite{liu2021spatial}, DICNN\cite{zhu2017realtime} and trained VFIT\cite{shi2022video}, DICNN\cite{zhu2017realtime} and Liu\cite{liu2021spatial} at two distinct parameter levels, 9 million (large) and 0.5 million (small), on Vimeo-90k train dataset\cite{xue2019video}. 

As shown in Table~\ref{tab:2}, our large model, \textit{Ours-L}, achieves state-of-the-art performance on all datasets and is the most efficient in terms of runtime and parameters. Moreover, our small model, \textit{Ours-S}, also performed the best among all of the small models with the least amount of parameters used. Relative to DICNN's original parameter count of 0.07M, increasing the model's parameter count to 0.5M led to improved performance. However, when the parameter count was further increased to 9M, due to the simplicity of the network architecture, it may have resulted in a loss of robustness leading to unsatisfying results. 
Therefore, we have excluded it from the comparison.

\begin{table*}[ht]
  \centering
  \begin{tabular}{l c c c c c}
    \hline
    \hline
           & Parameters(M)& VimeoTest &  Vid4 &  SPMC & UDM10  \\ 
    \hline
   w/o Image Alignment & 0.50 &  44.09 & 34.05  &  45.83&43.99\\
   Unidirectional Propagation   &0.50   & 43.35  & 33.37  &  44.13&44.50\\
   w/o FGDA             & 0.53 &  41.24 & 32.76  & 41.66 &42.71 \\
   Conv-ReLU Block           &  0.57     &  43.07      &  33.30      &   43.94    &  43.76     \\ \hline
   Our complete model & 0.50 &  44.40 & 34.20  & 46.35 &44.49\\
   \hline
   \hline
  \end{tabular}
\caption{Ablation study of the components. In each dataset, we evaluate in terms of PSNR. We conducted an ablation study on a small model (0.5M) across different datasets. To eliminate the influence of the reduced parameter count due to the absence of a specific component, we readjusted the network parameters to ensure they were all at the same parameter level, in order to ensure a fair validation of the effectiveness of each individual component.}
\label{tab:3}
\end{table*}

\subsection{Qualitative Results}
\label{sec:visual result}
\begin{figure*}[h]
    \centering
    \includegraphics[width=0.991\linewidth]{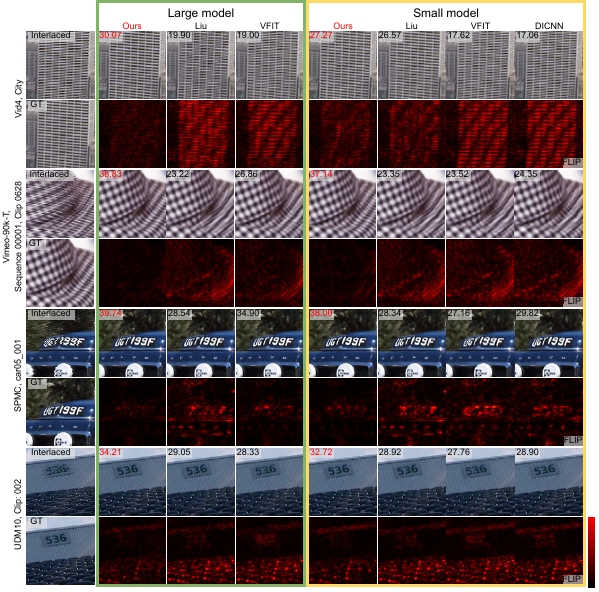}
    \caption{Visual comparisons of our method with existing deinterlacing methods. The first column shows the interlaced image and ground truth. The columns marked by green and yellow rectangles represent the results and FLIP\cite{Andersson2020} error maps from the large and small models, respectively. The PSNR values written in the top-left corner are computed for each crop. As depicted in the first and second scenes, the existing methods struggle to restore the distortion of the high-frequency repetitive patterns, while our method aligns with the ground truth. As demonstrated in the third and fourth scenes, our method achieves better fidelity on sequences with rapid camera movement.
    }
    \label{fig:vr_all}
\end{figure*}

\begin{figure*}[h]
    \centering
    \includegraphics[width=1\linewidth]{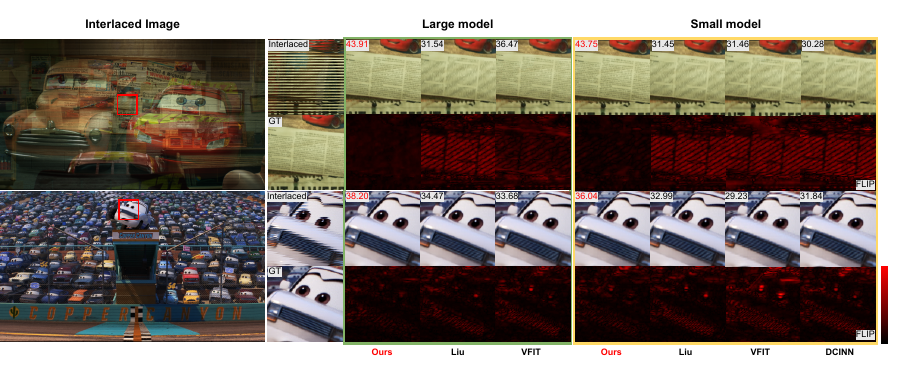}
    \caption{Visual comparisons showcase the deinterlacing results for animation content. Our method correctly restores the detail of the poster on the wall and the "nose" (intake grille) of the animated character. The PSNR values in the top-left corner are computed for each cropped region.}
    \label{fig:car}
\end{figure*}

In Fig.~\ref{fig:vr_all}, we present qualitative comparisons between our approach and alternative methods. To intuitively demonstrate the discrepancy between the models' prediction and the ground truth, we visualize the pixel level FLIP~\cite{Andersson2020} error maps where the brighter regions indicate more visible differences by human perception.
While other approaches also have succeeded in eliminating interlaced artifacts, they often fail to handle areas with intricate textures and details. 
Notably, our approach consistently produces sharper results across various datasets and reduces combing and aliasing artifacts when generating deinterlaced frames compared with existing methods. As shown in Fig.~\ref{fig:car}, our method can be generalized to animation content and consistently achieves surpassing performance in removing aliasing artifacts.

\begin{figure*}[h]
  \centering
  \includegraphics[width=1\linewidth]{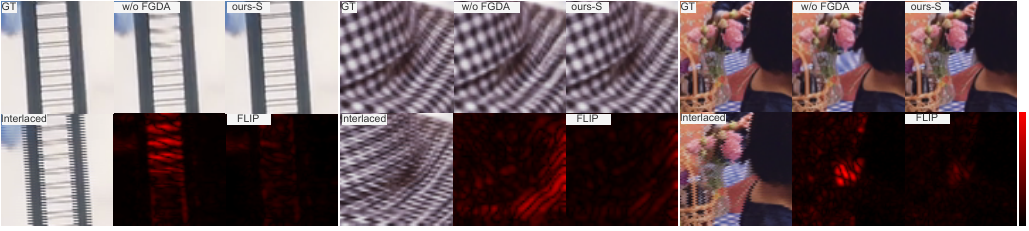}
  \caption{Visual results showcase the impact of the FGDA module in the ablation study. With the aid of the FGDA module, complex details are restored and aliasing artifacts are significantly alleviated.}
  \label{fig:ablation}
\end{figure*}

\section{Ablation study}

We devised several ablation studies to reason on our design and assessed the significance of each component within our network.

\textbf{Impact of Image-level alignment.}
In our proposed method, the fields that enter the network undergo image-level alignment before proceeding to latent-level alignment, propagation, and aggregation. To attest to the necessity of temporal alignment in color space, we removed the Image-level alignment, which resulted in a slight decline across all test sets, named \textit{w/o Image Alignment} in Table~\ref{tab:3}.

\textbf{Impact of Bidirectional propagation.} To motivate our bidirectional propagation approach to enlarge the temporal receptive field, we conducted a variant of our model utilizing only unidirectional propagation, labeled as \textit{Unidirectional Propagation} in Table~\ref{tab:3}.

The results demonstrate that models with unidirectional propagation produced varying degrees of performance reduction across different datasets due to the imbalance in aggregating temporal information. Note that the unidirectional propagation model exhibits a relatively minor performance drop on the UDM10~\cite{yi2019progressive} dataset, which can be attributed to the limited scale of motions in this dataset. In contrast, our complete model with bidirectional propagation gathers additional information from neighboring fields, resulting in enriched feature alignment, effectively preserving more details.

\textbf{Impact of FGDA module.}
The effectiveness of feature alignment in the temporal domain has been thoroughly analyzed in \cite{chan2022basicvsr++}. To ensure the completeness of our work, we removed all the FGDA modules so that the receptive field is constrained within individual fields, and the quantitative results are shown in \textit{w/o FGDA} in Table~\ref{tab:3}. Furthermore, as illustrated in Fig.~\ref{fig:ablation}, the significance of FGDA and bidirectional propagation scheme becomes more pronounced in regions that contain fine details and intricate textures. On one hand, in these specific regions, the available information from the current field is quite limited for the reconstruction process. Utilizing the bidirectional propagation scheme allows for the information to be transmitted through a robust and efficient propagation process. Essentially, this supplementary information aids in the restoration of intricate details. On the other hand, optical flows provide reasonable base measures for the deformable convolutions (DCN) and DCNs enhance the diversity of the optical flow, thereby enabling the offsets to capture more sophisticated temporal correspondence in highly distorted regions. 

\textbf{Impact of S-NAF Block in FRB.}
To motivate our choice of S-NAF as basic blocks in the network, we substitute them with the conventional Conv-ReLU residual blocks, as shown in \textit{Conv-ReLU Block} in Table~\ref{tab:3}. Our model with S-NAF offers a lighter architecture and improved performance.

\section{Conclusion}
In this paper, we introduce a novel deep learning-based video deinterlacing framework. To the best of our knowledge, our model is the first deep learning-based deinterlacing framework that takes into account both image and feature space bidirectional alignment in conjunction with feature refinement. To address the interlacing artifacts, we first employed a pre-trained SPyNet to obtain the forward and backward optical flows at four different scales. These flows have been used for field alignment in the image space and also later in latent space. For more accurate feature information propagation, we proposed a feature refinement Block (FRB), performing bidirectional propagation and refinement across different scales to expand the receptive field while effectively enhancing the utilization of temporal information. In the reconstruction process, we employed a residual mechanism both in the latent space and image space, facilitating a more effective reconstruction of the deinterlaced image. Notably, our model was designed to be capable of concurrently processing six fields of interlaced images, which reduces the processing time significantly. Through our extensive experiments, we demonstrate that our proposed method achieves state-of-the-art results while also providing the potential for real-time deinterlacing applications.



\printbibliography                

\newpage

\end{document}